# Fault Detection Engine in Intelligent Predictive Analytics Platform for DCIM

Bodhisattwa Prasad Majumder[1*], Ayan Sengupta[1], Sajal Jain[1], Parikshit Bhaduri[2]
[1]Post Graduate Diploma of Business Analytics, IIM Calcutta, ISI Kolkata and IIT Kharagpur
[2]GreenField Software Private Limited
Email: {bodhisattwapm2017, ayans2017, sajalj2017}@email.iimcal.ac.in, parikshit.bhaduri@greenfieldsoft.com

*Abstract: With the advancement of huge data generation and data handling capability, Machine Learning and Probabilistic modelling enables an immense opportunity to employ predictive analytics platform in high security critical industries namely data centers, electricity grids, utilities, airport etc. where downtime minimization is one of the primary objectives. This paper proposes a novel, complete architecture of an intelligent predictive analytics platform, Fault Engine, for huge device network connected with electrical/information flow. Three unique modules, here proposed, seamlessly integrate with available technology stack of data handling and connect with middleware to produce online intelligent prediction in critical failure scenarios. The Markov Failure module predicts the severity of a failure along with survival probability of a device at any given instances. The Root Cause Analysis model indicates probable devices as potential root cause employing Bayesian probability assignment and topological sort. Finally, a community detection algorithm produces correlated clusters of device in terms of failure probability which will further narrow down the search space of finding route cause. The whole Engine has been tested with different size of network with simulated failure environments and shows its potential to be scalable in real-time implementation.*

## I. INTRODUCTION

This year on a Monday morning Delta Airlines' operation came to a standstill [10]. Passengers check-in to the flights in London Heathrow were told that the flight check-in systems were not operational. The story was same across the globe during that time. Not only did this cause tremendous financial loss to Delta, it caused immense damage to Delta's brand. It turned out that the reason behind this outage was the failure of a critical power supply equipment, which caused a surge to the transformer and loss of power. The critical systems did not switch over to back up power. In other words, a power surge caused by single malfunctioning unit brought Delta's operation to a halt. One can find similar incidents caused by malfunctioning equipment across industries. Not only are data centers vulnerable due to sudden failure of equipment, so are other critical infrastructure such as electricity grids, utilities, airports etc.

Failures such as the one in Delta, leads to root cause analysis and instituting capabilities to predict such failures. The progress of machine learning and analytics has possibility to tap the vast amount of data from sensors, equipment, factories and machines not only to monitor the health of the equipment but also to predict when something is likely to malfunction or fail. With the advent of huge data availability, a predictive and prescriptive analytics platform is an inseparable part of any analytics software. It is crucial to understand the possibility of future events to take precautionary measure to prevent unwanted situation hence cutting the operational and maintenance cost. Data Center Infrastructure Management (DCIM) software, which monitors critical data center equipment, is also starting to utilize analytics capabilities to predict failures. At configuration stage, DCIM is mapped with the critical relationships and dependencies between all the assets, applications and business units in the data center. This makes it possible to identify cascading impacts of an impending failure. Over a period of time, data patterns evolve which lend themselves to modern predictive and prescriptive analytics. Predictive analytics gives the data center team enough time to take measures to either avoid or reduce the impact of the failure when it happens. [5] Players in the field of DCIM are advancing with their predictive technology giving health maps of devices, predicting downtime of a device based on past data, estimating the gap between forecasted values and real values in analysing alarms etc. [6, 7, 8].

---

* Corresponding author

It is imperative for data centers to keep uptime of the equipment to the maximum. This reduces the possibility of data center downtime. The various equipment in the data center such as UPS, PDUs, PACs are monitored by DCIM with real time alerts if critical parameters cross thresholds. The streaming data from devices as well as alerts and faults reports can be correlated by analytics to pin point root causes of failures as well as predict device failures. These aid immensely in streamlining data center operations.

The organization of the paper is as follows: Section II describes the complete objective of the research along with the problem definition. Section III discusses about related work and the novel contribution of this paper. Section IV elaborates the Proposed Fault Engine modules divided into several subsections. Section V presents the experimental setup and results. Section VI analyzes the time and space complexity of the proposed algorithms and finally section VII concludes the paper discussing its direct and derived business application tuned with scalability.

## II. PROBLEM STATEMENT AND OBJECTIVES

This paper contributes to a full stack design of predictive analytics platform in DCIM Software. The major contribution of this paper is as follows:

- The model of a computing unit, Fault Engine, which leverages the log-data from all concerned devices available in the device chain and employs a Markov Process based Failure Model to predict whether the failure is permanent or transient hence raising alarm with proper severity. It also indicates the recovery probability at any given time stamp after the failure has occurred.

- The Engine identifies the root cause devices in a situation of failure. It assigns probability to devices to be a probable root cause hence narrows down the search space for the DC management team.

- The Engine is capable of identifying communities of correlated devices based on the correlated failures.

## III. LITERATURE SURVEY

Though several previous work [8, 3, 17] discusses about the failure management in distributed system, but the integrated correlated alarm module with fault prediction is not reported in the literature yet, in our best knowledge. The Markov failure model have been reported in [18] with an assumption of exponential distribution which lacks to predict the time-dependent probability of recovery of a device at any instance after failure. The generalized assumption of Weibull distribution and its parameter estimation from available log data to enable the Fault Engine to predict the time dependent recovery probability at any time instance after the failure is indeed a meaningful extension of work [18]. Furthermore, the Bayesian probability assignment and topologically sorted list of probable root causes gives a unique flavor in terms of real-time implementation and inference making. The community detection algorithm have been discussed for several applications [19], but in our best knowledge, it never has been used in a network to cluster devices based correlated failure. The paper presents the complete implementation of Fault Engine in connection with middleware in DCIM software. The idea can be extended to any distributed system surveillance software to understand and predict the severity of a failure scenario. The complete model have been implemented for trial in the architecture of GFS Crane Software [11].

## IV. FAULT ENGINE

### IV.A. Permanent and Transient Failure Detection using Markov Failure Model

DCIM Software allows the Alarm module to raise alarms for individual device when it exceeds the already set threshold for parameters of the device but it does not guarantee that the failure is permanent while raising alarms. At times, it is costly to tackle false alarms which are not indicating transient failures. Hence, it demands a robust and intelligent alarm module which is able to raise alarm based on severity of failure i.e. persistent or temporary failure.

The fault engine conceptualizes a Markov Model, prevalent in distributed system control [18] which assumes three state for each device. Three states are Active (A), Transient Failure (T) and Permanent Failure (P) (Figure 1). It is intuitive and evident that a device can transit from Active state to Transient Failure state, and vice versa. Also, the device can transit from Active state to Permanent Failure state. Other transitions are not possible. To model the Markov Process, it is sufficient to estimate these three transition probabilities. We define these transition probabilities as follows:

- Probability of transition from A to T : $\alpha$
- Probability of transition from T to A : $\beta$
- Probability of transition from A to P : $\gamma$

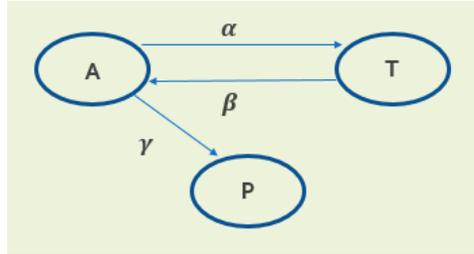

Fig. 1 State transition diagram for a node

These three parameters correspond respectively to the mean time to failure (*MTTF*) (given that the failure is transient), the mean time to recovery (*MTTR*), and the mean life time (*MLT*) of the node. Particularly, $\alpha = 1/MTTF, \beta = 1/MTTR, \gamma = 1/MLT$.

We define another variable called time to recovery (TTR) which defines number of time instances taken to be recovered from the failure. Hence $G(t)$ is the probability of a device being in the permanent failure state given $t$ consecutive instances of failure have been observed. That is, $Pr(TTR = \infty \mid TTR \geq t)$ is $G(t)$. We apply Bayes Theorem to derive the function G.

$$G(t) = Pr(TTR = \infty \mid TTR \geq t)$$
$$= \frac{Pr(TTR = \infty) \cdot Pr(TTR \geq t \mid TTR = \infty)}{Pr(TTR = \infty) \cdot Pr(TTR \geq t \mid TTR = \infty) + Pr(TTR \neq \infty) \cdot Pr(TTR \geq t \mid TTR \neq \infty)}$$
$$= \frac{Pr(TTR = \infty)}{Pr(TTR = \infty) + Pr(TTR \neq \infty) \cdot Pr(TTR \geq t \mid TTR \neq \infty)}$$

In calculation, $Pr(TTR = \infty)$ is the probability that a failure is a permanent failure. Since the failure transitions from state A has two competing transitions: one is to T with rate $\alpha$, and the other is to P with rate $\gamma$, $Pr(TTR = \infty) = \gamma/(\gamma + \alpha)$. The Markov model failure and recovery behaviours of a node as a stochastic process with transition periods between different states of the node distributed as Weibull distribution. Weibull distribution is the most generalised version of exponential family. It is assumed that as the time progresses, the probability of failure increases for a device and hence Weibull distribution is a suitable match to capture that time dependency. The parameters of the distribution $\lambda$ (scale) and $k$ (shape) can be estimated from the streaming data. Fig. 2 describes the distribution with different parameter values. Now, by substitution we get,

$$G(t) = \frac{\gamma}{\gamma + \alpha \cdot e^{-(\frac{t}{\lambda})^k}}$$

Furthermore, even if $G(t)$ indicates the probability of a device to become transiently failed, it does not indicate the probability of recovery at time t given it failed at time 0. The probability that a device will recover at time $t'$, given it has observed $t$ failures states ($t' > t$), can be obtained by the function $R(t, t')$, as below:

$$R(t, t') = \frac{Pr(TTR = t', TTR \geq t, t \geq t' | TTR \neq \infty)}{Pr(TTR \geq t | TTR \neq \infty)}$$

$$R(t, t') = \frac{(k/\lambda)(\frac{t'}{\lambda})^{k-1} \cdot e^{-(\frac{t'}{\lambda})^k}}{e^{-(\frac{t}{\lambda})^k}}, 0 < k < \infty, 0 < \lambda < \infty$$

The above expression comes from the similar analogy of hazard function in the study of survival analysis. Once we have obtained a device as transiently failed, we can further estimate the probability from the distributional assumption using the Weibull probability distribution function. This consolidates the fact why Weibull distribution has been used in estimating $G(t)$ because in the simplest form i.e. in exponential ($k = 1$), $R(t, t')$ will only depend upon the difference in $t$ and $t'$ which might not be the case. The memoryless property of exponential distribution fails to capture the information at time $t$.

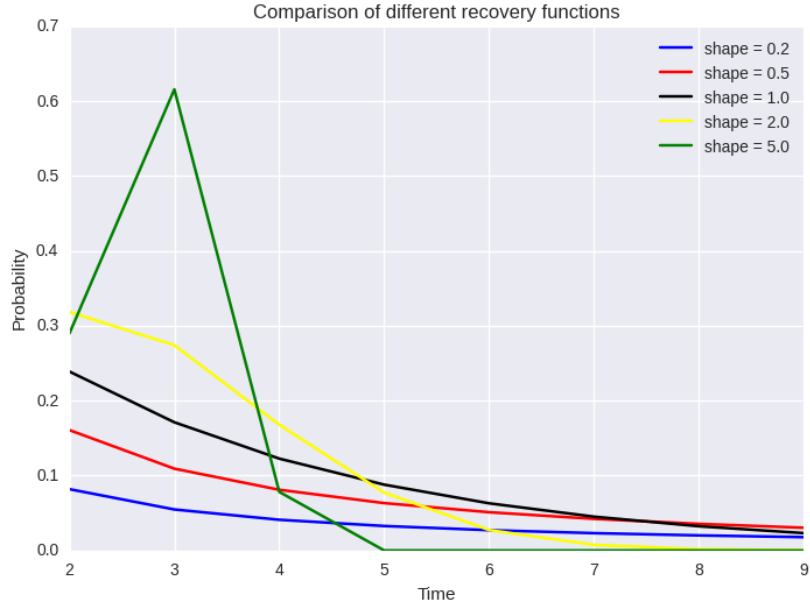

Fig 2. Plot of recovery function with different parameter estimates of Weibull function

For example, if the lifetime of a router is 6 time periods, and it has failed at 4th time periods, then the probability of recovery in 5$^{th}$ time period would be very less. This information has been captured by using the Weibull distribution, where the $R(t,t')$ does not only depend on the time difference between $t$ and $t'$. Hence, as the time progresses, the probability of recovery will decreases i.e. the shape parameter $k$ will be greater than 1. Fig. 2 indicates the asymptotical decreasing behaviour of recovery probability for $k$ values greater than 1.

The bottleneck of this method is the estimation of distribution parameters for Weibull distribution from streaming data because the whole stream (or a reasonably large sample) is needed every time for parameter estimation which poses a high space complexity. To do away with this problem, the assumption of Weibull distribution can be relaxed and the simplest form of exponential distribution ($k = 1$) can be assumed. Then $G(t)$ would look like:

$$G(t) = \frac{\gamma}{\gamma + \alpha.e^{-\beta t}}$$

The advantage of this model is that from an incoming data stream, it is convenient to calculate the mean ($\frac{1}{\alpha}, \frac{1}{\beta}, \frac{1}{\gamma}$) without storing the whole data stream in the memory hence space efficient. But the tradeoff has been made in estimating the function $R(t,t')$ which will be only dependent on the time difference of $t$ and $t'$. The pseudocode is in table Algorithm 1 with the exponential distributional assumption of recovery behavior. The idea extends the work in [18] employing a more generalized version of stochastic recovery behavior. Along with that, it introduces a novel function $R(t,t')$ from the analogy of hazard function which also estimate probability of recovery at an exact time point. The module has unique application in DCIM which helps to under the severity of failure and hence acts as an input into the next section of the engine. This module works in a local manner where device level probability has been calculated to detect the nature of failure which enables to generate alarm based on the severity of failure.

---

**Algorithm 1**

    procedure
2: *Prologue:*
        Calculate MTTR, MTTF, MLT from historical data.
4:    Calculate $\alpha = 1/\text{MTTR}, \beta = 1/\text{MTTF}, \gamma = 1/\text{MLT}$.
    *Input:*
6:    Input all the probable faulty device ids $\{d_1, d_2, \cdots d_n\}$ from the alarm module.
    *loop:*
8:    for $d_i$ do
        Input the current period of failure $t_{d_i}$ from alarm module.
10:    $G(t_{d_i}) = \frac{\gamma}{\gamma + \alpha e^{-\beta t_{d_i}}}$, $G(t)$ = Probabilty(failure of $d_i$ is persistent$|t$ consecutive failures have been observed)
        if $G(t_{d_i}) > threshold$ then
12:        failure is persistent.
        failure is probably transient.

---

**IV.B. Root Cause Analysis using Polling and Trap**

In an instance of failure, several devices can raise alarms depending on their status of availability irrespective of being the root cause of the failure. It is critical to identify the proper root cause, which minimizes operational cost and time of the DC Management team. DCIM software is usually capable of raising real time alarms on critical health devices but it is unable to justify the root cause of failure in a robust and synchronized manner [2]. The Fault Engine deploys the middleware in polling and trapping from all standard IP protocol (e.g. SNMP/MODBUS) enabled devices in every time instance set by the user in a synchronized manner [9]. A data center consists of several devices, mostly IP Protocol enabled, i.e. remotely accessible. Starting from the power station, all the devices create an acyclic chain or directed dependency graph where each device is a node and there is an edge between two nodes if they are physically/electrically connected with a direction towards the power/information flow. We assume the whole graph is acyclic i.e. there is no directed cycle in the graph. Also, in practical it is possible to have devices which is connected to dummy power supply to reduce possible link failure of its parent. Hence, we assume a node can have more than one parent. If one device fails, all of its dependent devices will eventually fail (except dummy parent cases). Fig. 3 describes a sample directed dependency network for different devices in the Data Center. We have considered all devices sending alarm as faulty as well as all the devices that are not reachable by polling-trap. Our goal of this module to allow the Fault Engine to know about the probability of a device to be actually faulty. We apply Bayesian rule to decide for device of being faulty given some its dependent is sending alarm or failed. Mathematically for devices $f_i$ and $d_j$,

$$Pr(f_i \text{ device is faulty} \mid d_j \text{ is sending alarm})$$
$$= \frac{Pr(d_j \text{ is sending alarm} \mid f_i \text{ device is faulty}) \cdot Pr(f_i \text{ device is faulty})}{\sum_{f_i} Pr(d_j \text{ is sending alarm} \mid f_i \text{ device is faulty}) \cdot Pr(f_i \text{ device is faulty})}$$

The marginal probability of failure for a device can be directly calculated from the historical alarm log simple by counting the number of failure for a device diving by total number time failure happened in the network. The conditional probability of sending alarm for a device given some other device is failed is an identity function $I_{f_i,d_j}$, i.e. $I_{f_i,d_j} = 0$ if $f_i$ is the parent of $d_j$, 0 otherwise. Thus we find out the conditional probability of being faulty for a device given all of its dependent devices sending alarm. All the probabilities for a particular device have been aggregated and divided by the total number of dependent devices are sending alarm and finally the mean conditional probability could be calculated for all faulty devices.

**Algorithm 2**
1: **procedure**
2: *Prologue:*
3:   $Dict \leftarrow \{\}$
4:   **for** $i$ in dependency graph $G$ **do**
5:     $count_i$ = number of times $i$ failed in historical data.
6:     $Dict[i] = count_i / \sum_{i=1}^{|G|} count_i$.
7: *Input:*
8:   Input all the alarming device ids $\{d_1, d_2, \cdots d_n\}$ from the alarm module.
9:   Master node sends poll to all the SNMP/MODBUS devices.
10:  $Q \leftarrow \emptyset$
11:  **for** $i$ in $G$ **do**
12:    **if** $i$ does not trap back **then**
13:      $Q.push(i)$.
14:    **if** $i$ traps back **Faulty then**
15:      $Q.push(i)$.
16: *main:*
17:  Do topological sort on $Q = \{f_1, f_2, \cdots f_m\}$.
18:  **for** $f_i$ **do**
19:    **for** $d_j$ **do**
20:      $P(f_i|d_j) \leftarrow \frac{Dict[d_j]I_{f_i,d_j}}{\sum_j Dict[d_j]I_{f_i,d_j}}$, where $I_{f_i,d_j} = 1$ if path $(f_i, d_j)$ in $G$ exists.
21:    $P(f_i) \leftarrow \sum_{d_j} P(f_i|d_j)/n'$  ; where $n'$ is total direct or indirect alarming children of $f_i$
22:    Output $P(f_i)$.

It is evident as we describe from the dependency structure and from the sample figure that a device resides in upper hierarchy is more critical as a root cause of failure than the ones reside in the lower part of the network. Hence, the Fault Engine is able to sort all the devices topologically from the acyclic directed dependency graph which gives a decreasing order of importance of devices. The engine further assigns the conditional probability calculated for each faulty device and a second level sorting had been done based on probability where the hierarchical level of two devices is same. The order given by this module narrows down the search space of devices to a collection of probable root cause. The pseudocode of the algorithm is given in Algorithm 2 table. The marginal and conditional probability update can be done in any instance with the use of streaming data without storing the whole data chain.

### IV.C. Root Cause Analysis with Correlated Alarm System

This extension of this previous module employs clustering between devices based on their correlated failures. The Fault Engine employs all the conditional probabilities given all other devices from the historical data to identify the correlation between devices based on their correlated failures. It is easy to understand the correlation in terms of failure among the nodes in one cluster hence the user will have information about the probability of other nodes from the same cluster of being the root cause.

An undirected graph have been constituted with all devices as nodes where the measure of conditional probability that probability of one node sends an alarm given the other node also sends an alarm, will act as an edge weight between these two nodes in question in this similarity graph of all devices. The engine operationalizes a community detection algorithm named Girvan-Newman method [16] in this new weighted similarity graph to identify the clusters or communities of devices which are highly correlated in a failure scenario.

**Algorithm 3**

1: **procedure**
2: *Prologue:*
3:     Take dependency graph $G = (V, E)$.
4:     **for** $i$ in $V$ **do**
5:         **for** $j$ in $V$ **do**
6:             Calculate $P(i|j) = \frac{Dict[j]I_{i,j}}{\sum_j Dict[j]I_{i,j}}$, where $I_{i,j} = 1$ if path $(i,j)$ in $G$ exists.
7:     Create similarity graph $G2 = (V2, E2)$, where $V2 = V$, $(i,j) \in E2$ if path $(i,j)$ exists in $G$ and $weight(i,j) = P(i|j)$.
8:     $Dict = \{\}$
9:     **for** $i \in V2$ **do**
10:         $Dict[i] = \sum_{j \in N(i)} weight(i,j)/|N(i)|$, where $N(i)$ is the neighbourhood of $i$ in $G2$.
11:     Cluster the nodes of $G2$ based on modularity optimization e.g. - Girvan-Newman community detection algorithm.
12: *Input:*
13:     Input all the alarming device ids $\{d_1, d_2, \cdots d_n\}$ from the alarm module.
14: *main:*
15:     $Q \leftarrow \emptyset$
16:     **for** $d_i$ **do**
17:         **for** $v_j \in \text{cluster}(d_i)$ **do**
18:             $Q.push(v_j)$.
19:     Do topological sort on $Q$.
20:     Sort $Q$ based on the $Dict$ value.
21:     $Q$ contains the probable faulty nodes in descending order of severity.

Girvan Newman is widely popular community detection algorithm which calculates the edge betweenness [12] of the graph and simply removes the edges having highest edge betweenness. It also optimizes the modularity function of the network given the cut as the objective function hence obtains fairly optimal clustering. Practically, the clustering algorithm targets to find clusters where one member raises alarm whenever another member from same cluster raises alarm. The advantage of this extension from previous module is the engine do not need to run the sorting over all the nodes. Whenever, in an instance of failure, a device alarms, the engine can deploy the sorting in between the cluster the alarming node belongs to. The clustering can be done offline or low-load scenario with a time interval and hence provides additional knowledge about the node failures during online. The fast implementation and integration of this module enhance the chance of identifying the right root cause very quickly when a failure happens. The Algorithm 3 follows the full description of this module.

**IV.D. Integration with Middleware – The Technology Stack**

It is important to seamlessly integrate all of the three modules and prepare a technology stack which sequentially performs the required task by three of these modules. Usually, all the log data are stored in Database which can be directly accessed by the middleware present in the system. Any DCIM software will possess an alarm module, be it separate or not, also can communicate with the middleware. We propose, the Fault Engine should be pushed in between the middleware and Alarm module. The Fault Engine doesn't require to directly access the databases due to security reason hence middleware can facilitate the information flow. Since Alarm module is just a record of all alarms in different faulty situations, Fault Engine needs to access the information. The marginal and conditional failure probability calculation requires the alarm log data along with parameter estimation also happens in the help of failure time stamp data. The polling-trap facility is available through middleware hence root-cause analysis can be done using proper information. The following diagram in Fig. 3 indicates the proposed stack design.

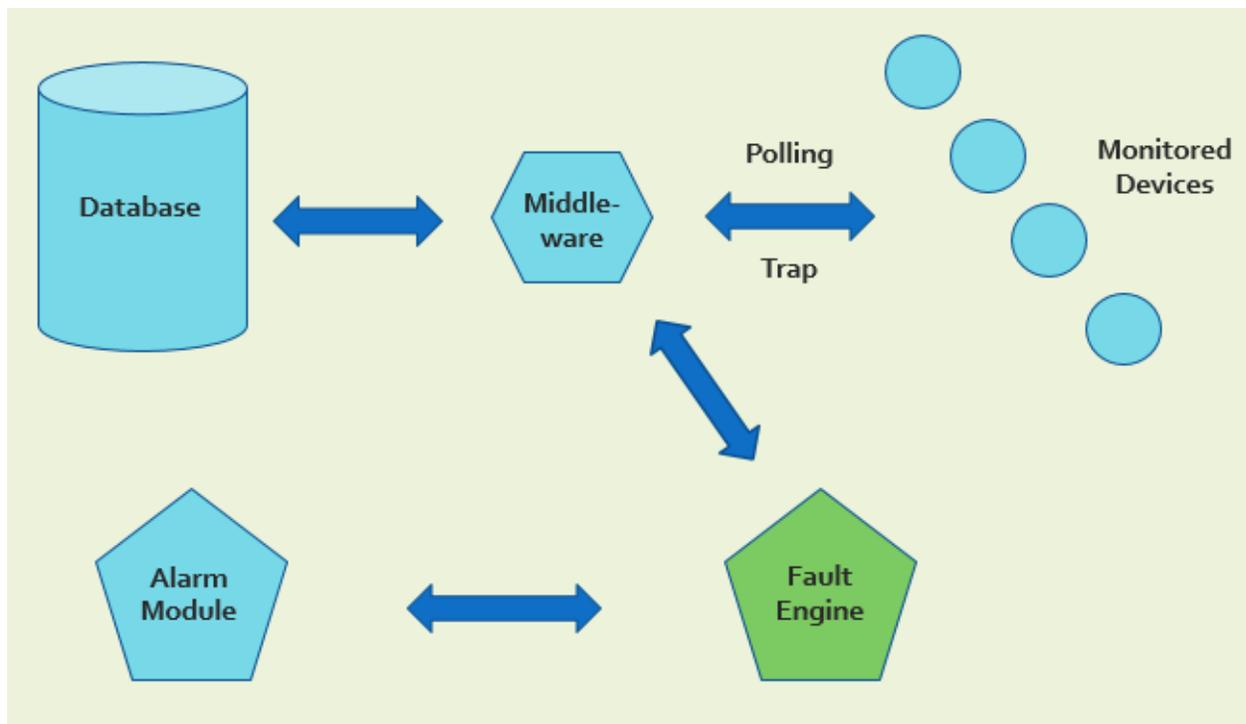

Fig. 3 Full stack implementation diagram

This concludes the operation of Fault Engine which sequentially runs the Markov model to determine the persistence of a failure in a local level, i.e. per device along with determining and then a global level graph clustering algorithm to detect clusters of device of correlated failure and then sorts all the devices inside a cluster topologically based on the directed dependence available in the DCIM power chain. The Fault Engine can be integrated between middleware and the alarm module to interact with both of them (Fig. 3).

## V. EXPERIMENTAL SETUP AND RESULTS

Most of the DCIM software do not store the data of actual root cause when they are detected with manual supervision [11]. The predictive system has not been included any of the DCIM software, as per our best knowledge. We have set up an experimental design with the technical expertise from Greenfield Software, a leading DCIM software company. We have developed a power chain consisting 47 devices acting as nodes. (Fig. 4). The network represents a generic power chain [11] in DCIM. The directed graph in the figure shows the hierarchy or precedence of the devices in order to be in chain. The higher order denotes more critical to the chain. We analyzed a case of PDU failure due to input power surge. The alarm logs have been generated for all devices below PDU2 according to real life case observed in GFS DCIM software. For all other devices, the overall error log to calculate the marginal error probability has been generated assuming exponential distribution. With practical consideration, we have assumed more than 25 failed log will indicate the permanent failure. Let us assume, 15 time stamps have been elapsed after the failure occurred in the PDU2 in the network of Fig. 4. Fig 5 shows the Weibull fit on the recovery time of PDU2 from its error log file. The estimated parameter have been reported as: $k = 2.43$, $\lambda = 4.69$. Similarly, the parameters of the Weibull distribution for all individual devices have been estimated from their respective sample error logs.

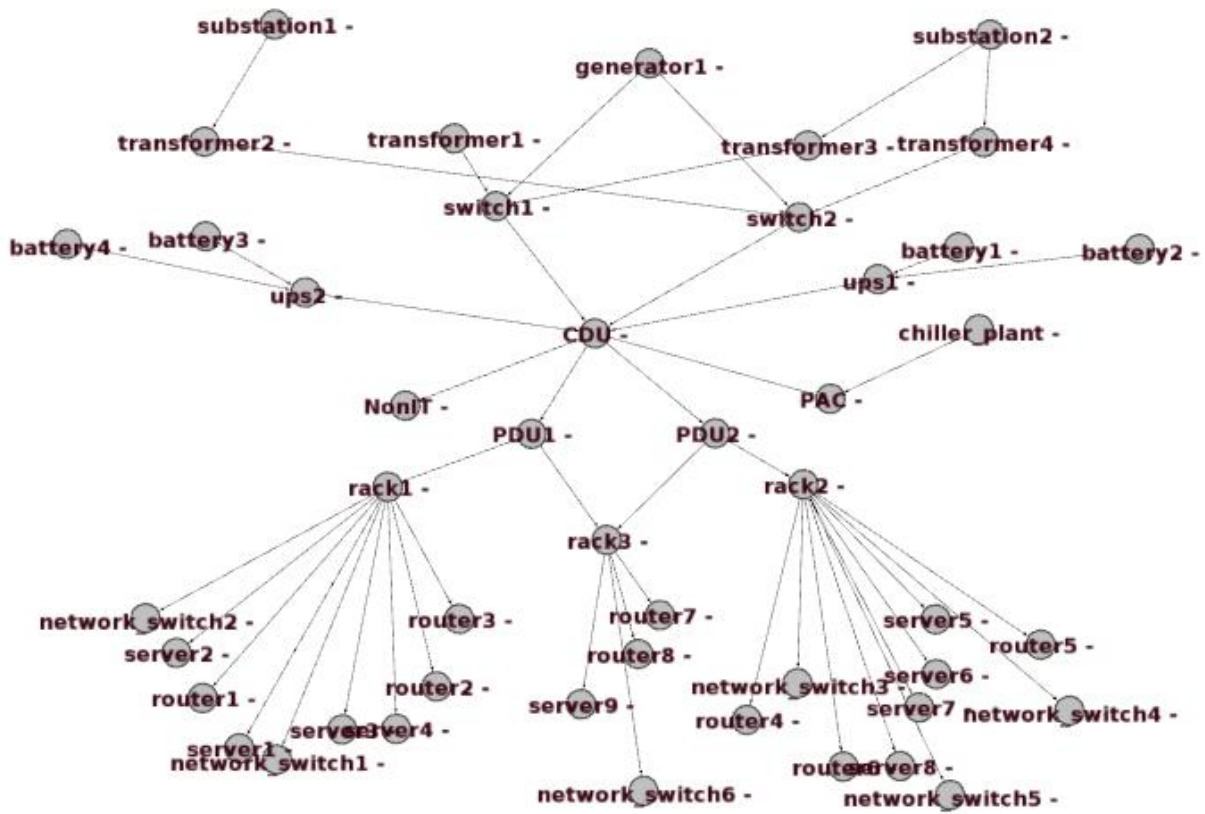

Fig 4. The dependency network of power chain (A generic constellation of 47 nodes)

Among PDU2 and its direct and children, PDU2, Rack2, Server6 and Router5 has showed non zero and high probability (0.8234, 0.7862, 0.7103 and 0.8165) of being permanently failure. We applied Root Cause analysis on these affected devices and the representative result shows particular racks (Rack2) has more probability of being the root cause than router in the given scenario as indicated by Fig. 6 (size of the node denotes its severity of failure). Also network has been segmented into 4 clusters based on their likelihood of being fail simultaneously or in correlated manner as indicated in Fig. 7. It is evident from the result that PDU2 and Rack2 falls in the same cluster. This clustering will certainly help in identifying related devices failure simultaneously.

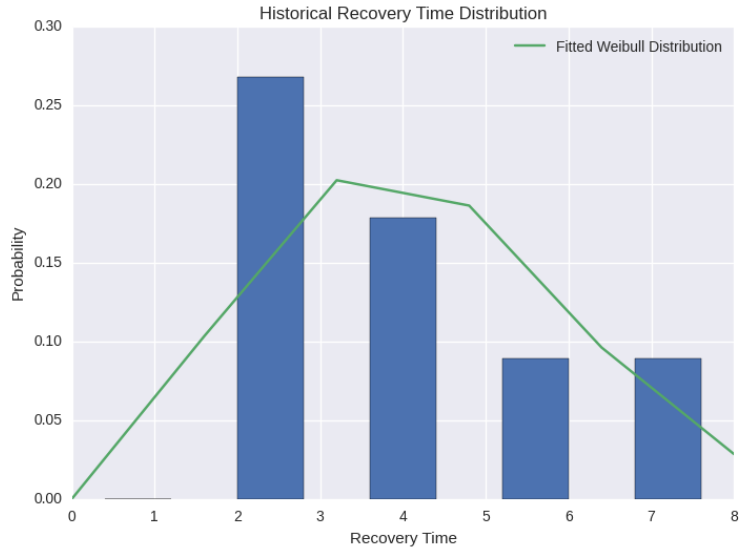

Fig 5. The Weibull distribution fit on recovery time data for PDU2

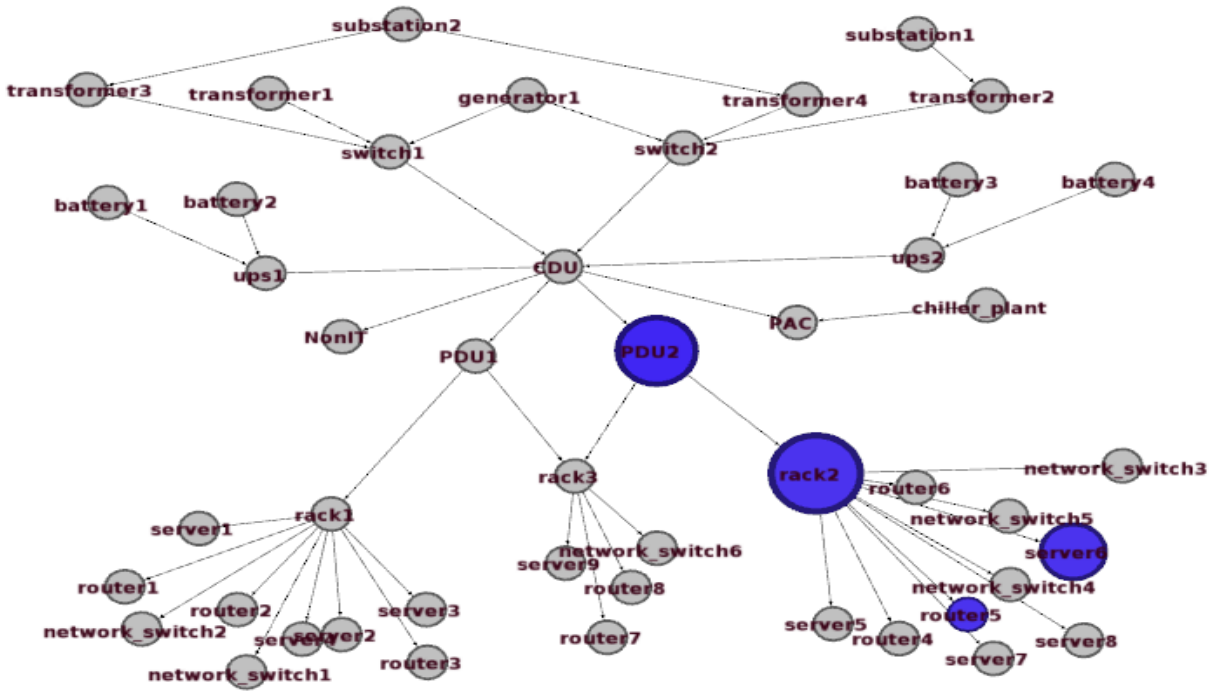

Fig.6. Scenario: PDU Failure for input power surge, Probable root causes are in blue color, size denotes the probability of being the actual root cause; the bigger, the more probable root cause.

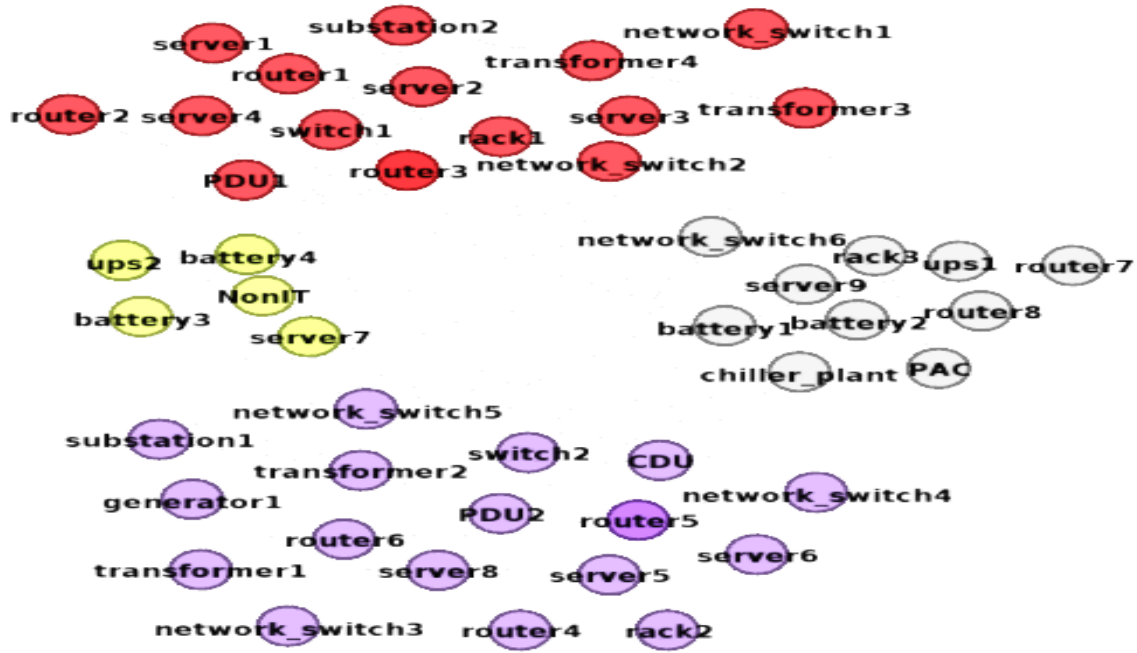

Fig. 7 Clusters of nodes on the basis of correlated failure, color denotes the cluster

The above experiment gave indication of the correctness of the result. To be certain about the scalability of the modules, we ran all of them sequentially on different size of the network having total nodes 50, 500, 5000, 12000, 50000, 100000. For all cases, the marginal error probabilities have been generated using exponential distribution with various values of the parameter. The next section reports the complexity and scalability of the modules based on this extended experiment.

**VI. TIME AND SPACE COMPLEXITY ANALYSIS**

We present a brief complexity analysis of all the algorithm which will indicate its applicability in DCIM software.

Time complexity of Markov Failure model for computing transient probability of a failed device is linear in the size of the system log file of the device. Hence, for each of the faulty devices (devices that are showing fault on their current log), the time complexity for fitting the Weibull distribution takes $O(K)$ time ($K$ is the size of the logfile which is equivalent to the time elapsed $t$), whereas, other constants like MTTF, MTTR and MLT can be computed in $O(1)$ from the streaming data. As the whole log file has been stored, the space complexity is also $O(K)$. Both time and space complexity can further be reduced to $O(1)$ with the use of streaming data, if we use non parametric failure models instead of any parametric distributional forms. But non parametric implementation cannot produce future failure rate predictions, which might be very crucial and vital for some practical scenarios. Fig. 8 shows the system and user runtime for the Markov Failure module which indicates almost constant over number of nodes in the network.

In the Root Cause Analysis module, the initialization takes $O(n)$ time to generate the graph with $m$ devices and $n$ edges. For a sparse graph like power/information chain dependency graph we often observe that $m \sim n$. Using the serial computation, topological sort takes $\Theta(n + m)$ time [13], which is asymptotically same as $\Theta(m)$. This complexity can be reduced to $\Theta(log^2 m)$ by using parallel computers of $O(m^k)$ number of processors. There are many implementations available viz. Floyd-Warshall [4], Thorup [15], Kameda's algorithm [14] for calculating all pair reachability in a network. Our implementation using Kameda's algorithm takes $O(m)$ preprocessing time as $O(1)$ time for each reachability query. The required calculation on the historical fault log takes $\Theta(K)$ time, where $K$ is the size of the fault log; $K \in O(t * m)$, where $t$ is the time elapsed. Finally, the step for calculating Bayesian probability for each of the $f$ number of probable faulty devices for each of the $a$ number of alarming devices takes $O(f * a * m)$ time. Accumulating all the steps, we compute the running time of our root cause calculation algorithm to be $O(m + t * m + f * a * m)$. Similarly, the space requirement can be calculated to be $O(m^2 + t * m)$. Fig. 9 indicates system runtime and user runtime taken with increasing number of devices in the network. As it shows, with 1 Lakh devices, the algorithm takes ~ 3 minutes to find the probable root causes. This can be reduced further if Root cause analysis is done on correlated cluster, where number of nodes in a cluster will be much less than the total size of the network. Even with 20,000 devices, the algorithm could output probable root causes within a minute.

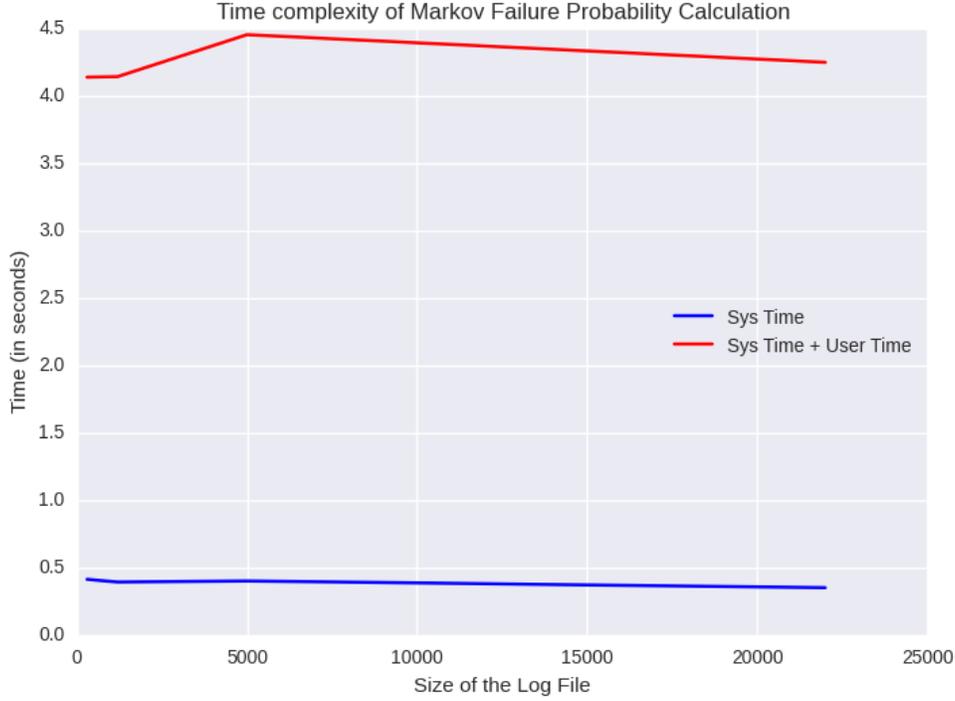

Fig. 8. Running time of Markov Failure Model

The algorithm finding correlated failure cluster takes $O(m^3)$ preprocessing time for calculating the undirected weighted fault dependence graph, as for each pair of vertices, we calculate the Bayesian probability of failure. The Girvan-Newman community detection algorithm [16] takes $O(n * m^2)$ time to produce the community structure among the devices. This time heavy process can be substituted by some faster implementations namely Clauset-Newman-Moore [1], which takes $O(m * log^2 m)$ time for a sparse graph with $m$ nodes. Hence, combining these two steps, the clustering implementation takes $O(m^3)$ time to produce the cluster of correlated devices based on their Bayesian failure probability for the network graph with $m$ devices. Fig. 10 shows the user runtime almost follows a cubic curve with increasing number of nodes in network. Particularly this algorithm doesn't store any additional value hence the space complexity for this algorithm is nothing though the space taken by the graph is $O(m^2)$.

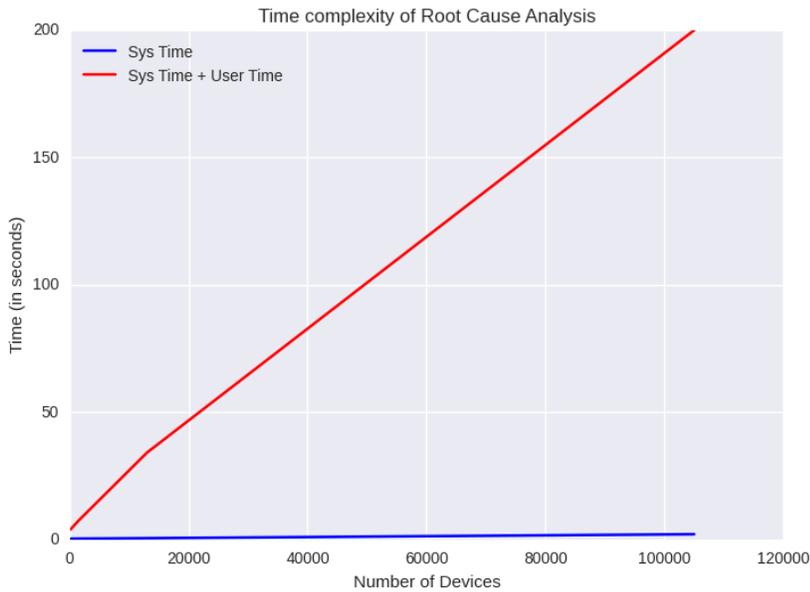

Fig. 9. Running time of Root Cause Analysis Module

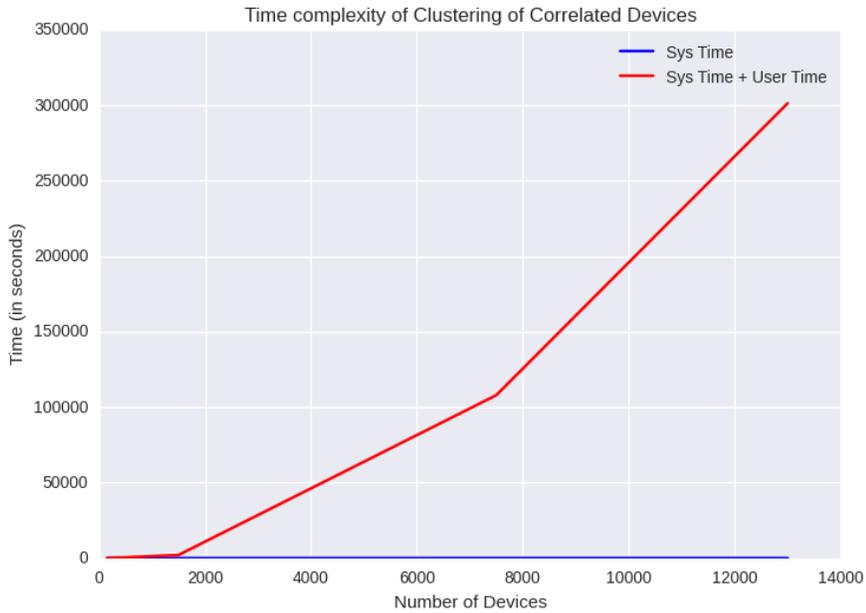

Fig. 10. Running time Root Cause Analysis module with Correlated Alarm System

**VII. BUSINESS IMPORTANCE AND CONCLUSION**

Delta's outage and similar outages across critical infrastructure have reinforced the need to predict failure before it happens. Hence, businesses are turning to predictive analytics to better understand equipment failure pattern so that they can take necessary steps to prevent outage. Apart from scalability of the Fault Engine, it provides and array of direct and indirect business importance when closely used with DCIM or any surveillance software for distributed system. The whole design is adjustable to integrate in any similar type software system like DCIM and hence produces immense applicability immediately. Some of the indicating business importance have been presented here.

- The predictive analytics platform will help customer to continuously monitor all the devices, to take precautionary actions based on the output probability which will save huge operating cost of recovering from the acute failure scenario.

- The predictive system will allow the user to understand the severity of the failure of a device and based that short-term or long-term action could be taken.

- The recovery function allows the user to understand the possibility of recovery of a failed device and based on the requirement the DC management can take back-up options in case of delayed recovery.

- The interactive platform will save human time and labor by making all the predictions automatic. It will significantly reduce the human error because of the leads to probable correct outputs.

- Alarm module is one of the most critical modules in DCIM. Correlated Alarm system will help customer to identify correlated devices which will lead to a guided search for actual root cause of the failure.

- The prediction for the severity of the failure will give space for the DC management team to understand the cause of failure beforehand and take subjective decisions before it worsens.

- Overall, the automated predictive analytics platform will give a global view of all devices as well local insights which in turn will help the customers to analyze and take meaningful actions.

A faster implementation of the whole technology can be done as a future extension of this work. The utility of this Fault Engine has been discussed in the context of a large data center. Similar module can be implemented in any distributed system for monitoring and control, bringing the benefits of failure prediction and thereby reducing business loss.